\documentclass[10pt,twocolumn,letterpaper]{article}
\usepackage[accsupp]{axessibility}
\usepackage{cvpr}              

\usepackage{graphicx}
\usepackage{amsmath}
\usepackage{amssymb}
\usepackage{booktabs}
\usepackage{algorithm}
\usepackage{algorithmic}
\usepackage{widetext}
\usepackage{multirow}
\usepackage{stfloats}
\usepackage{booktabs}
\usepackage{makecell}
\usepackage{mathtools, cuted}

\usepackage[pagebackref,breaklinks,colorlinks]{hyperref}

\usepackage[capitalize]{cleveref}
\crefname{section}{Sec.}{Secs.}
\Crefname{section}{Section}{Sections}
\Crefname{table}{Table}{Tables}
\crefname{table}{Tab.}{Tabs.}

\def\cvprPaperID{7} 
\def\confName{CVPR}
\def\confYear{2022}

\begin{document}

\title{Transfer Learning from Synthetic In-vitro Soybean Pods Dataset\\
for In-situ Segmentation of On-branch Soybean Pods}
\author{ Si Yang\textsuperscript{1,2,3} , Lihua Zheng\textsuperscript{1,2}, Xieyuanli Chen\textsuperscript{3}, Laura Zabawa\textsuperscript{3}, Man Zhang\textsuperscript{1,2}, Minjuan Wang\textsuperscript{1,2,*} \\
\textsuperscript{1} College of Information and Electrical Engineering, China Agricultural University \\
\textsuperscript{2} Key Lab of Smart Agriculture Systems, Ministry of Education, China Agricultural University \\
\textsuperscript{3} Institute of Geodesy and Geoinformation, University of Bonn, Germany \\
{\tt\small \{yangsi, zhenglh, cauzm, minjuan\}@cau.edu.cn, \{xieyuanli.chen, zabawa\}@igg.uni-bonn.de}
}

\maketitle


\begin{abstract}
The mature soybean plants are of complex architecture with pods frequently touching each other, posing a challenge for in-situ segmentation of on-branch soybean pods. Deep learning-based methods can achieve accurate training and strong generalization capabilities, but it demands massive labeled data, which is often a limitation, especially for agricultural applications. As lacking the labeled data to train an in-situ segmentation model for on-branch soybean pods, we propose a transfer learning from synthetic in-vitro soybean pods. First, we present a novel automated image generation method to rapidly generate a synthetic in-vitro soybean pods dataset with plenty of annotated samples. The in-vitro soybean pods samples are overlapped to simulate the frequently physically touching of on-branch soybean pods. Then, we design a two-step transfer learning. In the first step, we finetune an instance segmentation network pretrained by a source domain (MS COCO dataset) with a synthetic target domain (in-vitro soybean pods dataset). In the second step, transferring from simulation to reality is performed by finetuning on a few real-world mature soybean plant samples. The experimental results show the effectiveness of the proposed two-step transfer learning method, such that AP$_{50}$ was 0.80 for the real-world mature soybean plant test dataset, which is higher than that of direct adaptation and its AP$_{50}$ was 0.77. Furthermore, the visualizations of in-situ segmentation results of on-branch soybean pods show that our method performs better than other methods, especially when soybean pods overlap densely. 
 
  \textbf{Keywords:}Deep learning; Transfer learning; Computer vision; Instance segmentation; Plant phenotyping
\end{abstract}

\section{Introduction}
The crop yield of soybean (Glycine max L.) is heavily influenced by three major factors: the number of pods per plant, the number of seeds per pod and the seed size \cite{fehr1988principles,uzal2018seed}. The most crucial parameter is hereby the number of pods per plant, which is an important agronomic indicator \cite{mcguire2021high}. Identifying and segmenting on-branch soybean pods is the prerequisite for acquiring morphological phenotypic traits of mature soybean plants. Traditionally, soybean pod counting is performed manually. However, the soybean pods are small  with various shapes and the uncertain number of pods per plant as well as random occlusion by branches and other pods. This results in an error prone, time-consuming, and labor-intensive manual phenotyping procedure~\cite{riera2020deep}. Therefore, it is infeasible for large-scale mature soybean plant phenotype investigation.

\begin{figure*}[tp]
\begin{center}
\includegraphics[width=0.9\linewidth]{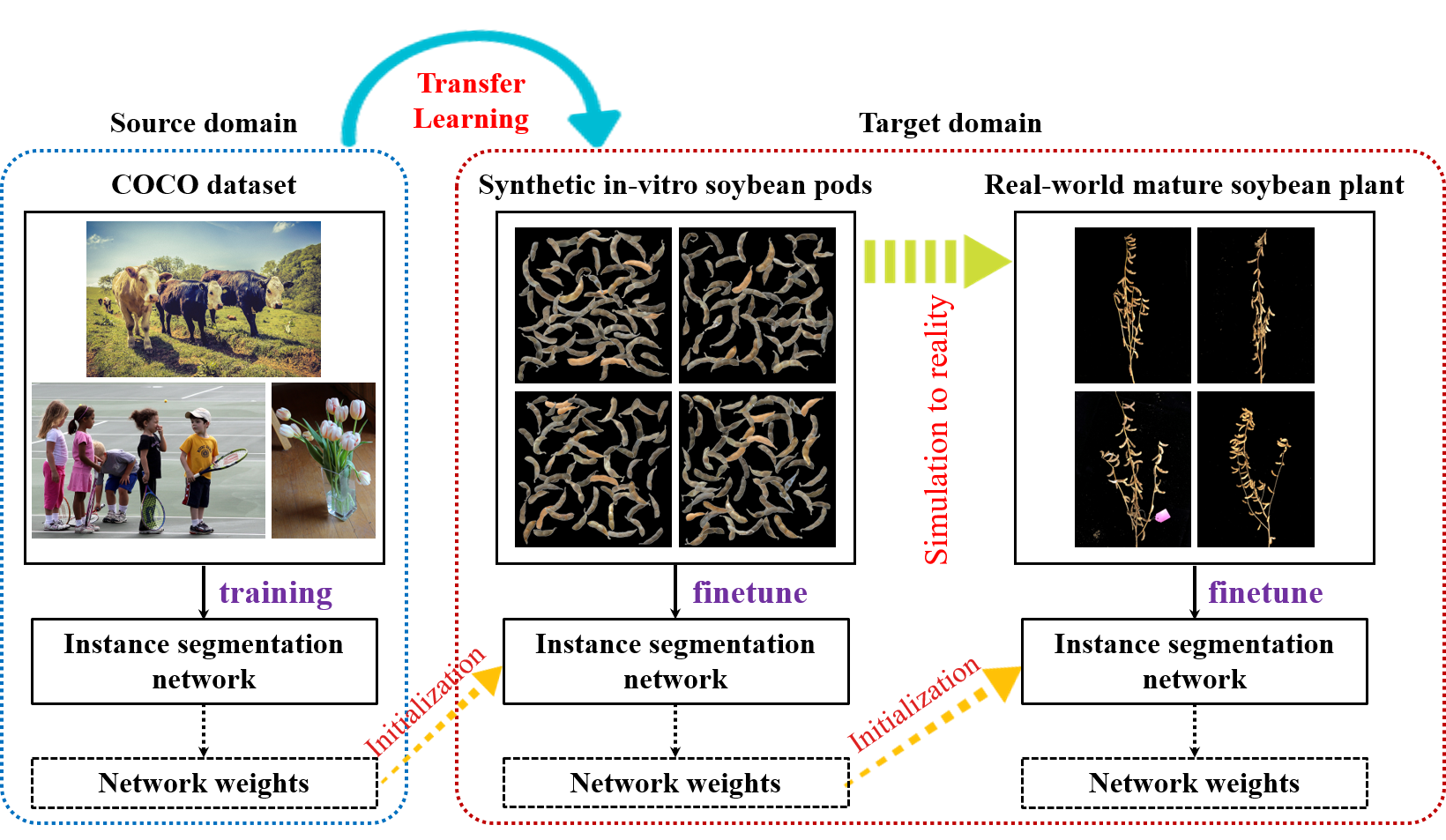}
\end{center}
\vspace{-0.5cm}
   \caption{Overview of the proposed two-step transfer learning for in-situ segmentation of on-branch soybean pods}
\label{fig:1}
\end{figure*}

Aided by the rapid gains in imaging technology, high throughput phenotyping became possible when the crops are sparsely or regularly placed with weakly physical contact \cite{baek2020high,yu2019fruit}. Some open-source image analysis methods have been used in high throughput crop phenotyping. The main idea of these methods is segmenting crops based on classic and ordinary image processing techniques, such as the watershed algorithm \cite{lamprecht2007cellprofiler,igathinathane2008shape}, morphological opening and closing operation \cite{tanabata2012smartgrain}, and tailored algorithm with handcrafted features \cite{groves2010estimating, yang2020efficient}, etc. Nevertheless, these methods can realize high throughput crop instance segmentation, but they are sensitive to changing illumination conditions and texture features of object, and are also inadequate in robustness and generalization ability. Additionally, the crops need to be sparsely distributed with small overlap under a consistent light condition to ease segmentation.

Deep learning has greatly improved the performance  on almost every computer vision task, and can achieve an effective segmentation by learning deep features from large annotated image datasets to solve the above-mentioned problems \cite{liu2021swin}. Some existing methods are used in high throughput crop phenotyping, including fruit counting, detection, and instance segmentation \cite{zabawa2020counting,pound2017deep,nellithimaru2019rols,yu2019fruit}, where the changes in the shapes of fruits are relatively small. Deep learning applied in plant phenotyping has grown exponentially in the past few years \cite{kamilaris2018deep}. However, training an accurate deep learning algorithm with strong generalization ability requires a large amount of labeled data which is one of the disadvantages of deep learning. Compared with relatively common tasks, such as classification in the ImageNet dataset\cite{deng2009imagenet} and object detection in COCO dataset\cite{lin2014microsoft}, the 
demands for large annotated data for specialized tasks in agricultural applications is even more pronounced \cite{desai2019automatic,ghosal2019weakly,chandra2020active}. However, manual annotation is expensive, especially for the instance segmentation tasks in the plant phenotyping realm. Although many techniques aim to decrease the cost of manual labeling, such as domain adaptation \cite{sakurai2018two} or active learning \cite{chandra2020active}, without compromising performance, the tedious, painful, labor-intensive, and time-consuming labeling process is still needed to evaluate the algorithm. Especially in high-throughput crop phenotyping, the annotation of crop instance dataset poses a tremendous challenge.

An option to reduce the cost of manual annotation is learning from synthetic data \cite{dwibedi2017cut,ubbens2018use}. Although the synthetic data is not faithful compared with real-word images, the critical characteristic  of synthetic dataset is that ground truth annotations can be automatically obtained~\cite{barth2018data}. Furthermore, the synthesizing data approach is able to create almost unlimited amount of labeled datasets~\cite{danielczuk2019segmenting}. Furthermore, synthetic data can represent changes in a variety of conditions, which is usually difficult to achieve through applying image augmentation techniques on real world images \cite{xu2018srda}. Kuznichov et al. \cite{kuznichov2019data} proposed a method to segment and count the leaves of Arabidopsis, avocado, and banana by using synthetic leaf textures with different sizes and angles to simulate images obtained in real agricultural scenes. Toda et al. \cite{toda2020training} proved that a synthetic dataset, rendering the combination and direction of seeds, was sufficient to train an instance segmentation network to segment barley seeds from real-world images. Collectively, synthetic datasets have great potential in the computer vision-based plant phenotyping research field \cite{yang2022synthetic}.

On-branch soybean pods segmentation and counting pose a challenge, since they feature a complex architecture and a high level of overlap  with each other. As lacking the labeled data for training an in-situ segmentation model of on-branch soybean pods, we present a two-step transfer learning method based on a synthetic high throughput in-vitro soybean pods dataset which is prepared by our novel automated image generation method. Figure \ref{fig:1} illustrates the overview of our proposed two-step transfer learning for in-situ segmentation of on-branch soybean pods.

The main contribution of this study are as follows:

(1) A novel synthetic image generation method is proposed  for automatically creating  labeled high throughput in-vitro soybean pods image sets.

(2) A new hybrid sim/real and in-vitro/on-branch dataset, including synthetic in-vitro soybean pods and real-world mature soybean plants, is designed for transferring from simulation to reality and from in-vitro segmentation to on-branch segmentation robustly. 

(3) The proposed two-step transfer learning method with a tiny Swin transformer-based instance segmentation network achieves a decent performance for in-situ segmentation of on-branch soybean pods. To the best of our knowledge, this is the first work utilizing synthetic high throughput in-vitro soybean pods for simulating on-branch soybean pods of mature soybean plants.

\section{Methods}

\subsection{Synthetic in-vitro soybean pods image generation}
\begin{figure}[t]
\begin{center}
\includegraphics[width=1\linewidth]{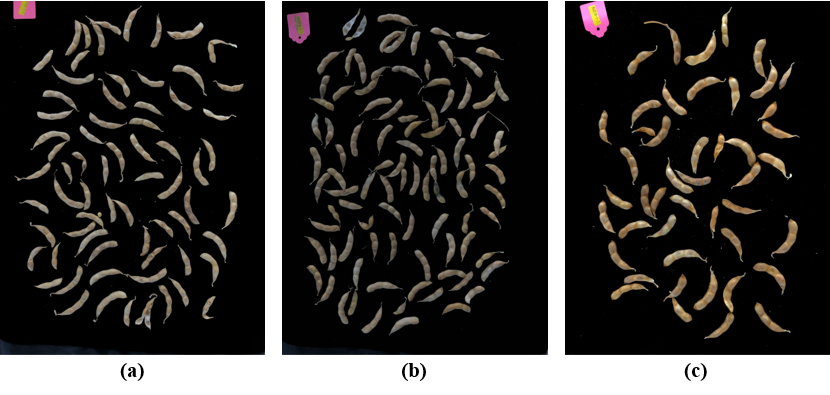}
\end{center}
\vspace{-0.5cm} 
   \caption{Examples of scanned raw in-vitro soybean pods images in which the soybean pods were picked from one plant sample manually and randomly tiled upon the black-colored flannel}
\label{fig:2}
\end{figure}

In our proposed pipeline of computer vision-based high throughput in-vitro soybean pods phenotype investigation, the soybean pods are picked manually from a single plant sample. They are then tiled upon a simple background like black-colored flannel randomly \cite{chang2021exploring}. After that, these pods are scanned by the camera sensor of an iPhone 8 plus (Apple) mounted on a tripod and saved as an image with the size of $3024\times4032$ shown in Figure \ref{fig:2}. The working distance of the camera sensor was fixed at about 30\,cm above the background.

\begin{figure}
\begin{center}
\includegraphics[width=1\linewidth]{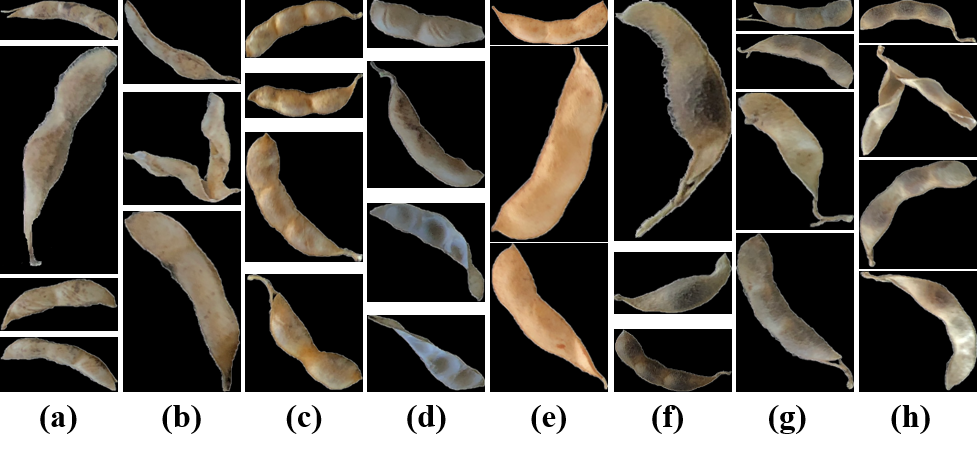}
\end{center}
\vspace{-0.5cm} 
   \caption{Examples of SPIP. each column represent a different cultivars. (a) BJ103, (b) BJ125, (c) BJ218, (d) BJ226, (e) BJ264, (f) BJ335, (g) BJ351, (h)BJ356}
\label{fig:4}
\end{figure}

 At the stage of dataset preparation,  we randomly chose 6 samples for each soybean plant accession (total of 48; 6 samples for 8 cultivars). One sample of each soybean plant accession (total of 8 soybean plants; one sample for 8 cultivars) is chosen  to create the synthetic image dataset. The soybean pods picked from one sample manually are  tiled above the black flannel randomly, scanned by our camera sensor, and saved as an individual image file. The rest samples of each accession (total of 40 soybean plants; 5 samples for 8 cultivars) are used as real-world on-branch soybean plants training dataset.

We now introduce our proposed method for generating synthetic images. First, we prepare a background image pool (BIP) and a soybean pod image pool (SPIP). The BIP was consist of 10 images, which  prepared by capturing the actual black flannel background 10 times and cropped at the fixed size of $1024\times1024$. The SPIP was constituted by segmenting each pod from raw in-vitro high throughput soybean pods images and saved as an individual pod image processed with Photoshop$\footnote{Adobe Photoshop CC 2020, Adobe Systems Incorporated, San Jose, CA, USA; http://www.adobe.com/Photoshop}$, as shown in Figure \ref{fig:4}. The images of the SPIP are zero-padded, to prevent the pod from moving outside the image after rotation and shift. Last, generate high-throughput in-vitro soybean pods images based on the BIP and SPIP prepared in the previous step. Algorithm~\ref{alg:1} depicts the detail of the above described last step, generating high-throughput in-vitro soybean pods image,  using pseudo code.



\begin{algorithm}[htb]
\caption{ Generation of fine-labeled high throughput in-vitro soybean pods image} 
\label{alg:1} 
\begin{algorithmic}[1] 
\REQUIRE  BIP and SPIP

\ENSURE  a high-throughput in-vitro soybean pods image, a fine-labeled mask image with different color
\STATE  create a raw canvas, a mask canvas.
\STATE  paste a background image chosen from the BIP randomly onto the raw canvas.
\STATE  fill the mask canvas with absolute black color.

\WHILE {generate a satisfying coordinate or less than max iteration}
\STATE  { select one pod image from SPIP randomly.}
\STATE  { rotate, shift and zoom at random.}
\STATE { generate new pod position coordinate $\left(x_i, y_i\right)$  randomly.}
\IF {$\left(0, 0\right) < \left(x_i, y_i\right) < \left(w, h\right)$ and overlap degree  of $\left(x_i, y_i\right)$ and $\left(x_j, y_j\right)$ satisfy $threshold$ }
\STATE  { paste the selected pod image onto the generated coordinate of canvas.}
\STATE  { create the mask image with different color according to the selected image.}
\STATE  { paste the created mask image on the mask canvas’s counterpart position.}
\ENDIF
\ENDWHILE 
\end{algorithmic}{}
\end{algorithm}

Note that, in Algorithm \ref{alg:1}, $\left(x_i, y_i\right)$ is coordinate of the $i^{th}$  pod will be pasted on canvas, while $\left(x_j, y_j\right)$ is coordinate of $0^{th} \sim j^{th}$  pods has been pasted on canvas. $w$ and $h$ are the width and height of canvas. The parameter $threshold$ is dynamic calculated related to the size of the soybean pod bounding box, in detail, the width of $i^{th}$ and $j^{th}$ soybean pod. The new coordinate  $\left(x_i, y_i\right)$  is generated randomly but with two limitations, the boundary of the canvas as well as the minimum Euclidean distance between the newly generated coordination  $\left(x_i, y_i\right)$   and the coordinates  $\left(x_j, y_j\right)$   of the soybean pods pasted on the canvas before. These two restriction factors keep the pod within the canvas, and adjust the degree of overlap respectively. 




\subsection{Real-world mature soybean plant dataset}
\begin{figure}[t]
\begin{center}
\includegraphics[width=0.8\linewidth]{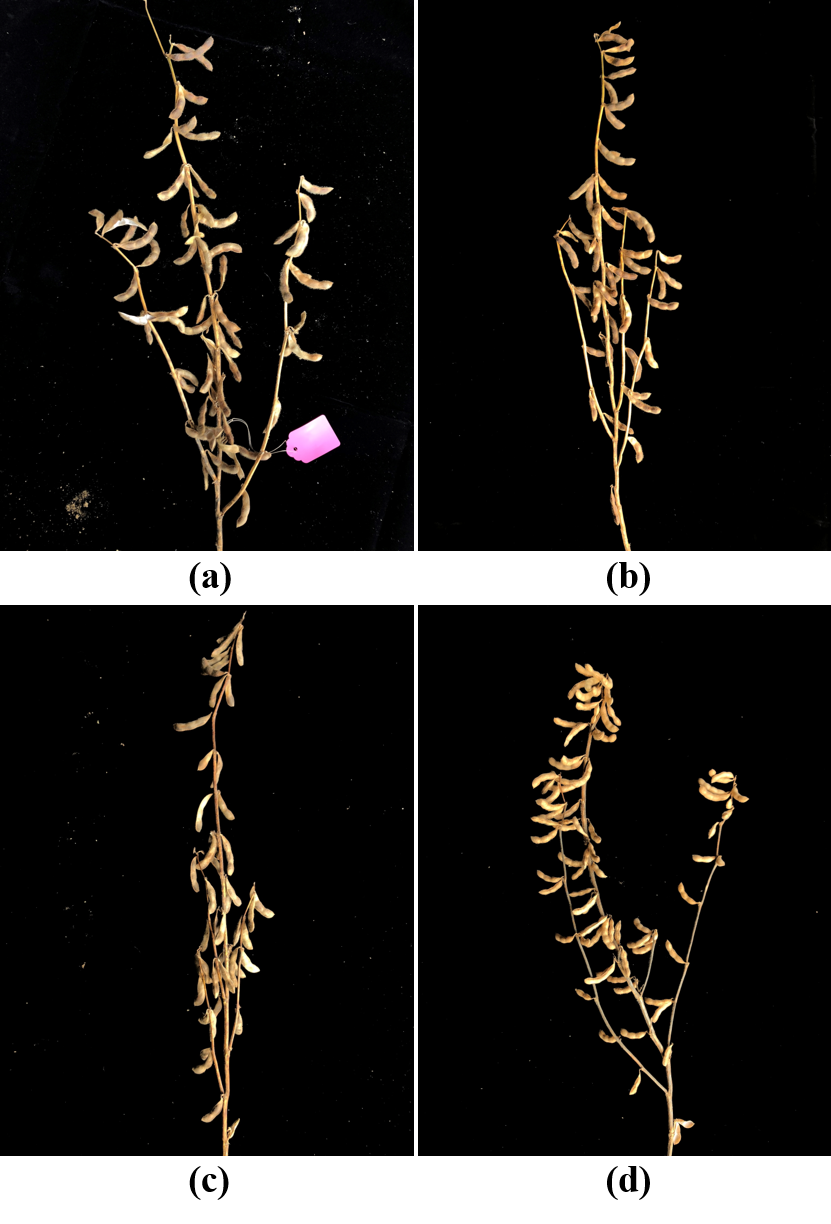}
\end{center}
\vspace{-0.5cm} 
   \caption{ Examples of the captured mature soybean plant images which the background is  black-colored flannel}
\label{fig:5}
\end{figure}

As described in section 2.1, we generate the synthetic labeled high throughput in-vitro soybean pods images. Now we are preparing a real-world mature soybean plant dataset by capturing images of individual mature soybean plants using the camera sensor of an iPhone 8 plus (Apple) mounted on a tripod, and saving them as an image with the size of $3024\times4032$ as shown in Figure \ref{fig:5}. The working distance between the camera sensor and the black-colored flannel background is adjusted to about 1\,m to capture the whole soybean plant image. We randomly chose 60 specimens of mature soybean plants to constitute the real-world mature soybean plant dataset. 40 of them are used for training and validation mentioned in section 2.1, 20 for testing. In addition, the mature soybean plants are  not used in generating the synthetic in-vitro soybean pods image dataset. After manual annotation, the images are down-scaled since the size of the image acquired by our came sensor is $3024\times4032$, which is too large for the computing of the instance segmentation network.

\subsection{Model training for two-step transfer learning}

\begin{figure*}[t]
\begin{center}
\includegraphics[width=0.9\linewidth]{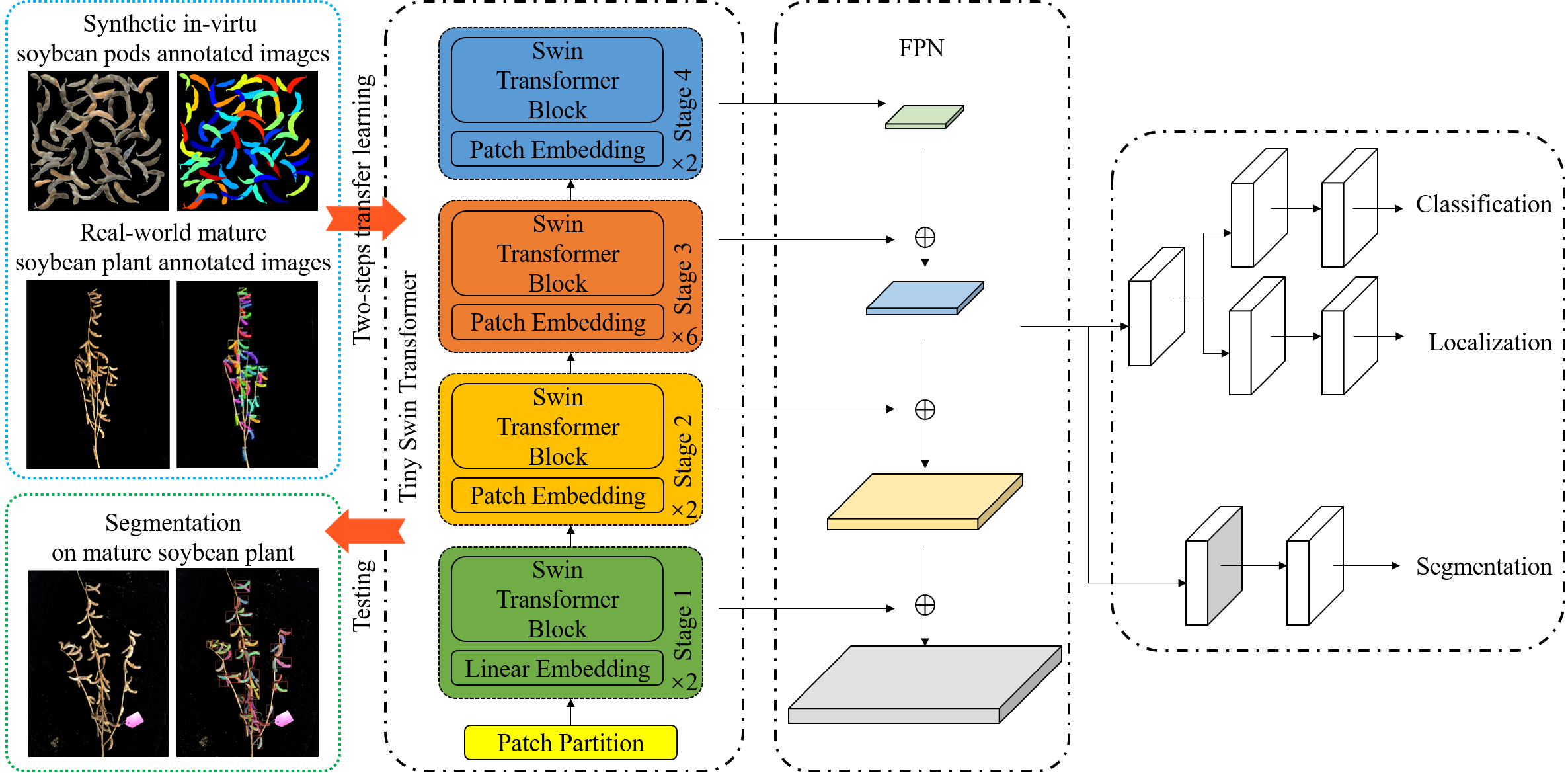}
\end{center}
\vspace{-0.5cm} 
   \caption{Tiny Swin Transformer based on-branch soybean pods instance segmentation of mature soybean plant with two step transfer learning}
\label{fig:6}
\end{figure*}

Swin Transformer, a hierarchical Transformer whose representation is computed with shifted windows, performed pretty well in dense prediction tasks \cite{liu2021swin}. To adapt the tiny Swin Transformer to our in-situ segmentation of on-branch soybean pods of mature soybean plants with complicated structures and frequent physical overlap, we reduce the number of classes to two. Each instance mask is classified as a foreground object or background, while only visualizing the foreground mask. We use our synthetic in-vitro soybean pods dataset with large amounts of labeled data and a real-world mature soybean plant dataset with a few labeled samples for our two-step transfer learning in the target domain. In the first step, we finetune the pre-trained model weights trained by the MS-COCO dataset \cite{lin2014microsoft} on our synthetic in-vitro soybean pods dataset. Then, in the second step, transferring from simulation to reality is performed by finetuning on the few real-world mature soybean plant samples, as shown in Figure \ref{fig:6}. To increase the diversity of dataset, the up-down, rotation, brightness and Gaussian blur image augmentations are used herein. The mean pixel value is set as the average pixel value of the simulated dataset. Tiny Swin Transformer is chosen as the backbone to exact features. Two evaluation metrics included average precision (AP) and recall, used to evaluate in the original research \cite{he2017mask}, are also 
introduced herein as the evaluation criteria. The comparative experiments of different instance segmentation network with our two step transfer learning is supplied in the supplemental material.

\section{Experiments and results}

\subsection{Software libraries and hardware}
The processing unit was a computer with an Intel Core i5-10400F@2.90Hz CPU, 16GB RAM, and a single GPU (8G, Geforce GTX2060 supper, NVIDIA). The synthetic image generation-related processes operated on the environment, including Integration Develop Environment (IDE), integrating Python 3.6, and OpenCV3 (ver. 3.4.2). The manual annotation of real-world mature soybean plant images was implemented with the Photoshop program.

\subsection{Preparation of pods instance segmenatation dataset}
We generated high throughput in-vitro soybean pods images with the size of $1024\times1024$. The pods were randomly and densely located inside the canvas region by our procedure, as mentioned in section 2.1.

\begin{figure}[t]
\begin{center}
\includegraphics[width=1\linewidth]{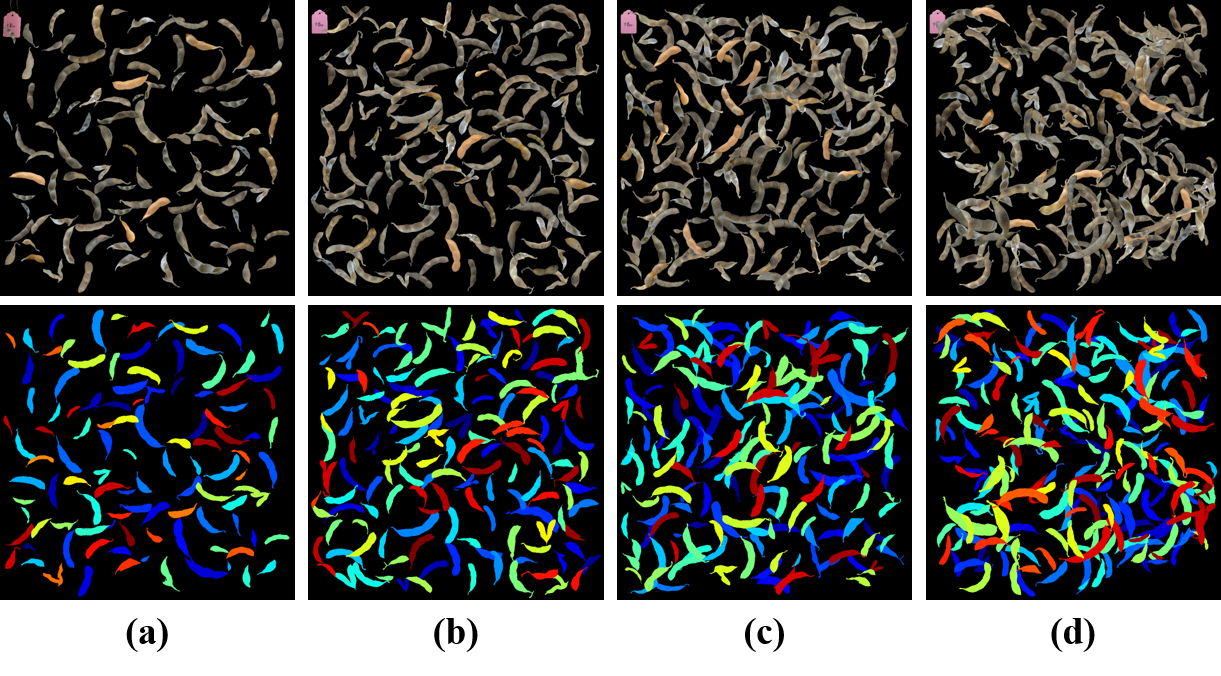}
\end{center}
\vspace{-0.5cm} 
   \caption{The synthetic in-vitro high throughput soybean pods image with different overlapping degrees. (a) the overlapping coefficient is 0.4 (b) the overlapping coefficient is 0.3 (c) the overlapping coefficient is 0.2 (d) the overlapping coefficient is 0.1}
\label{fig:fig8n}
\end{figure}

We prepared a number of training datasets of synthetic in-vitro soybean pods images to do transfer learning from the source domain (MS COCO dataset) to the target domain (synthetic in-vitro soybean pods). The prepared synthetic datasets include different amounts of (training/validation/testing) data with different overlapping degrees as shown in Table \ref{tab:table 1 } and Table \ref{tab:table 2 } to validate the efficiency of data mount and overlapping degrees of in-vitro soybean pods for in-situ segmentation of on-branch soybean pods. In other words, we investigated 40 datasets, 10 datasets for all 4 different overlap degrees. The visualization of different overlapping degrees of soybean pods is shown in Figure \ref{fig:fig8n}. We also prepared another new 200 sets of image pairs as a synthetic test dataset, and those synthetic images were not used in the model training or validation. The overlapping degree is a decisive coefficient for the purpose of calculating the dynamic overlap threshold as mentioned in section 2.1. When the overlapping degree decreases, the overlap threshold also decreases, resulting in a dense overlap of soybean pods. The overlapping degree is the key parameter to simulate the real world frequently physically touching on-branch soybean pods in our proposed synthetic dataset generation scheme.

\begin{figure}[t]
\begin{center}
\includegraphics[width=0.9\linewidth]{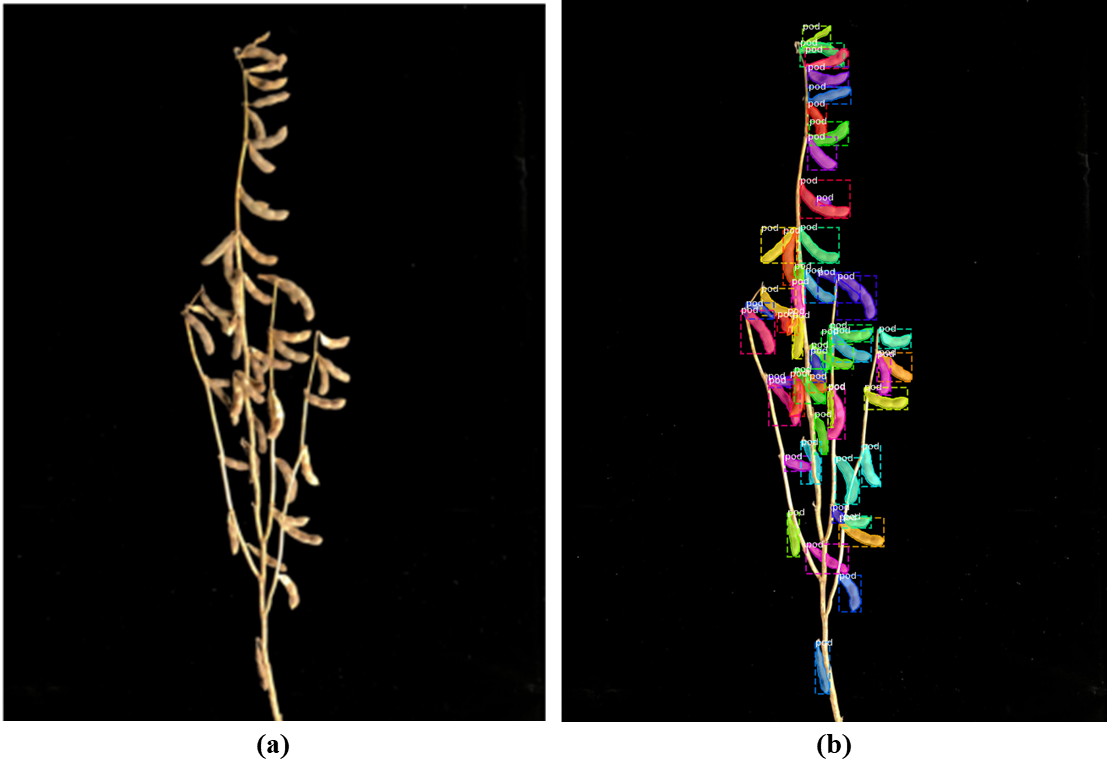}
\end{center}
\vspace{-0.5cm} 
   \caption{One labeled image data of real-world soybean pods test dataset annotated by Adobe Photoshop software. (a) raw image, (b) instance masks, bounding boxes, class names}
\label{fig:8}
\end{figure}

An example of the real-world mature soybean plant dataset is shown in Figure \ref{fig:8}. The real-world mature soybean plant dataset is labeled by Photoshop. The software is used for pod matting, and each matting pod was saved as a binary image. We found that there is a plethora of soybean pods per image leading to more time intensive labeling. The time of manual annotation process with LabelMe is about 90 min per image. The real-world mature soybean plant dataset is constituted of 60 image pairs of raw soybean images and its mask images, 36 of those images for training, 4 for validation, and 20 for testing.

Overall, the prepared different amounts of datasets for our two-step transfer learning consist of an in-vitro soybean pods dataset (Dataset\_in-vitro) and a real-world mature soybean plant dataset (Dataset\_on-branch) summarized in Table \ref{tab:table 3 }. The Dataset\_in-vitro is constituted of synthetic in-vitro soybean pods images and real-world in-vitro soybean pods images.


 \begin{table}[t]
\centering
 \caption{ Synthetic in-vitro soybean pods datasets of one overlap degree with different amounts of (training/validation/testing) data.}
\begin{tabular}{cccc}
\toprule[1.5pt] \text { Dataset ID } & \text { Training } & \text { Validation } & \text { Testing } \\
\midrule[0.6pt] 1 & 200 & 20 & \multirow{10}{*} {200} \\
 2 & 400 & 40 \\
 3 & 600 & 60 \\
 4 & 800 & 80 \\
 5 & 1000 & 100 \\
 6 & 1200 & 120 \\
 7 & 1400 & 140 \\
 8 & 1600 & 160 \\
 9 & 1800 & 180 \\
 10 & 2000 & 200 \\
\toprule[1.5pt]
\end{tabular}
\label{tab:table 1 }
\end{table}

 \begin{table}[t]
\centering
 \caption{ Synthetic in-vitro soybean pods datasets with different overlapping coefficients.}
\begin{tabular}{cccc}
\toprule[1.5pt] \makecell{Overlapping \\ coefficient}  & \makecell{ Image  \\ size } &  \makecell{Pod \\  count}  &  \makecell{Processing \\ time per image/s}  \\
\midrule[0.5pt] 0.1 & \multirow{4}{*} {1024 * 1024} & 212 & 28 \\
 0.2 &  & 192 & 25 \\
 0.3 &  & 156 & 22 \\
 0.4 &  & 112 & 18\\
\toprule[1.5pt]
\end{tabular}
\label{tab:table 2 }
\end{table}
 
 \begin{table*}[t]
\centering
 \caption{The detail of the prepared dataset for our two-step transfer learning}
\begin{tabular}{ccccccc}
\toprule[1.5pt]  &  \multicolumn{3}{c}{\begin{tabular}{c}
\text { Synthetic in-vitro soybean pods dataset } \\
\text { Dataset in-vitro }
\end{tabular}} &  \multicolumn{3}{c}{ \begin{tabular}{c}
\text { Real-world mature soybean plant dataset } \\
\text { (Dataset on-branch) }
\end{tabular}} \\
\midrule[0.5pt] & \text { Training } & \text { validation } & \text { Testing } & \text { Training } & \text { validation } & \text { Testing } \\
\midrule[0.5pt] 
\multicolumn{3}{c}{\makecell{Different amount of (training/validation) data \\ as  Table \ref{tab:table 1 } illustrated }} & 200 & 36 & 4 & 20 \\
\toprule[1.5pt]
\end{tabular}
\label{tab:table 3 }
\end{table*}


\subsection{Efficiency of different amounts of data with different overlapping degree}
\begin{figure*}[t]
\begin{center}
\includegraphics[width=0.8\linewidth]{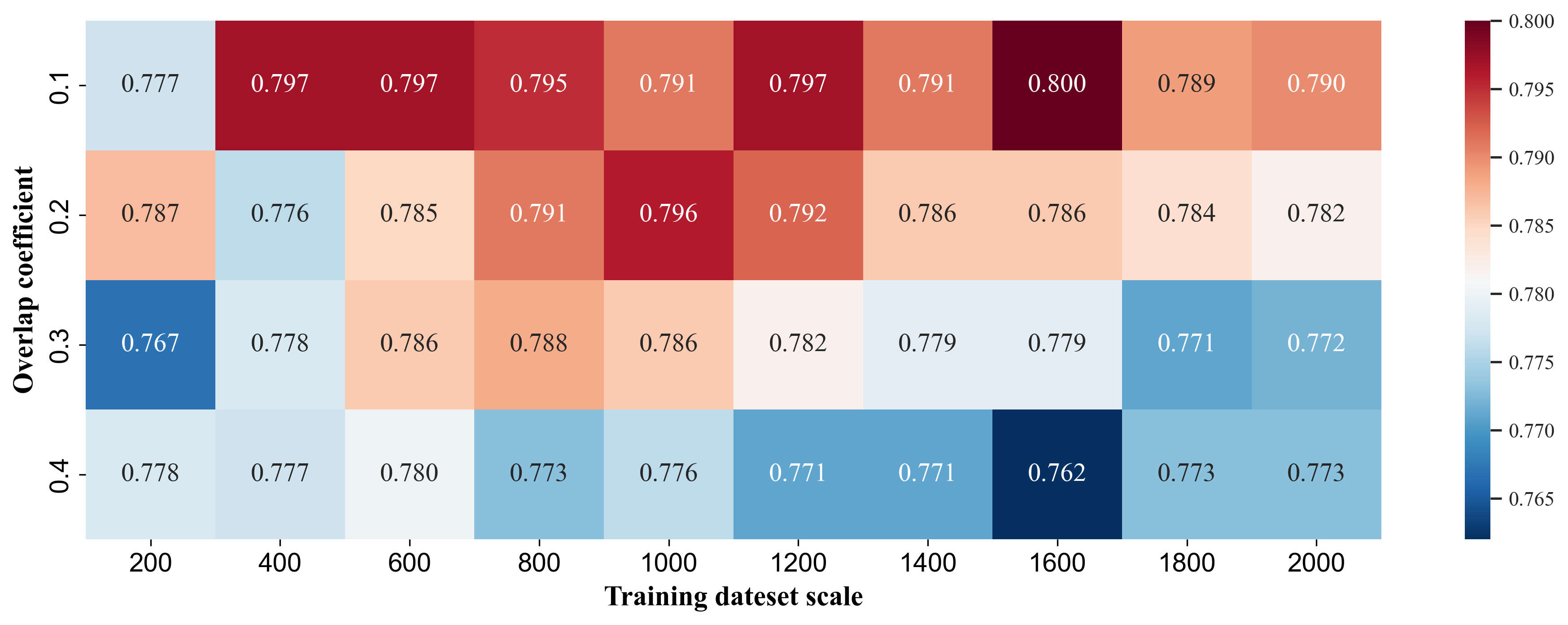}
\end{center}
\vspace{-0.5cm} 
   \caption{The heat map of evaluation metric (AP$_{50}$) instance segmentation results on real world mature tiny Swin transformer based two-step transfer learning model trained with different amounts of data with different overlapping coefficient.}
   \vspace{-0.3cm}
\label{fig:9}
\end{figure*}

To investigate the impact of different amounts of data with different overlapping degrees, we explicitly present a set of two-step transfer learning experiments on the prepared dataset as mentioned in section 3.2. We use our real world mature soybean plant test dataset to evaluate the tiny Swin transformer-based two-step transfer learning model trained on the different amounts of data with different overlapping degree datasets. The variation in AP$_{50}$ of different datasets is shown in Figure \ref{fig:9}. We 
found that the best performance is transfer learning the dataset which includes 1600 synthetic in-vitro soybean pods images and its overlapping degree is 0.1. And we can witness that increasing the overlapping degree can increase the performance dramatically. We also learned that increasing the training data amount is not significant.

\subsection{Effectiveness of two-step transfer learning}
We investigate the effectiveness of our two-step transfer learning from the synthetic in-vitro high throughput soybean pods dataset for in-situ segmentation of on-branch soybean pods. As mentioned in section 3.2, the prepared dataset for our two-step transfer learning consists of an in-vitro soybean pods dataset (Dataset\_in-vitro) and a real-world mature soybean plant dataset (Dataset\_on-branch). The Dataset\_in-vitro was constituted of synthetic in-vitro soybean pods images and real-world in-vitro soybean pods images.

(1) COCO, Dataset\_in-vitro. The backbone network parameters were initialized with the parameters trained by COCO dataset. Then the network is finetuned by our synthetic in-vitro soybean pods dataset (Dataset\_in-vitro) in the target domain.

(2) COCO, Dataset\_on-branch. The backbone network parameters were initialized with the parameters trained by COCO dataset. Then the network is finetuned by real-world mature soybean plant dataset with fewer samples (Dataset\_on-branch) in the target domain which is one-step transfer learning.

(3) COCO, Dataset\_in-vitro, Dataset\_on-branch. This is the proposed approach, named "Ours". The backbone network parameters were initialized with the parameters trained by COCO dataset. Then the network is finetuned by our synthetic in-vitro soybean pods dataset (Dataset\_in-vitro). We then apply transfer learning from Dataset\_in-vitro to Dataset\_on-branch.

\begin{figure*}[htp]
\begin{center}
\includegraphics[width=0.90\linewidth]{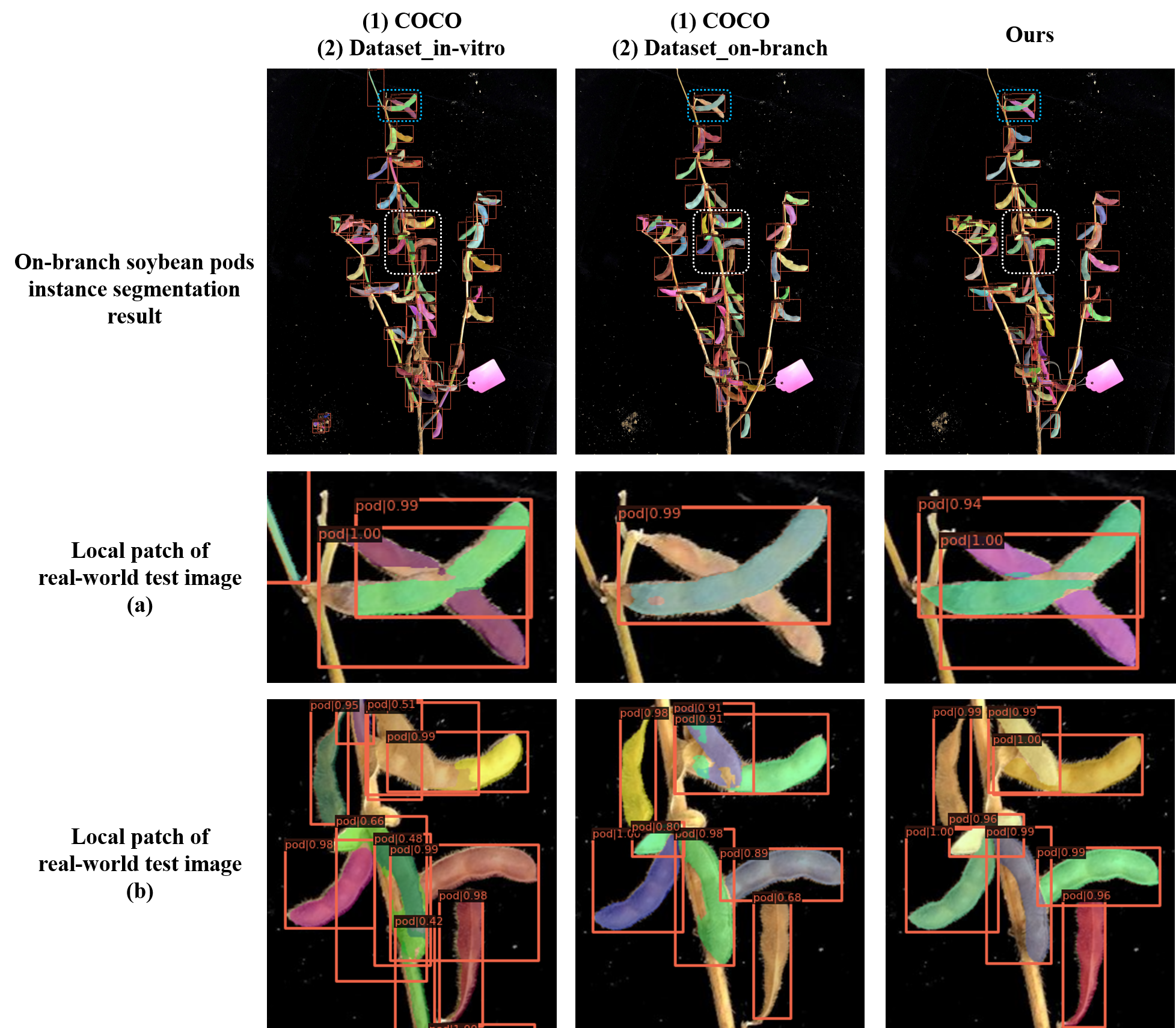}
\end{center}
\vspace{-0.5cm} 
   \caption{One example of visualized output results of real-world mature soybean plant dataset with different transfer learning strategies. The network is tiny Swin Transformer based instance segmentation network. The first column is the result of the pre-trained model purely retrained by our synthetic in-vitro soybean pods dataset. The second column is the result of the pre-trained model only retrained by our real world mature soybean plant. The third column is the result of the pre-trained model first retrained by our synthetic in-vitro soybean pods dataset and finetuned on our real world mature soybean plant dataset.}
\label{fig:10}
\vspace{-0.4cm} 
\end{figure*}

The quantitative and qualitative results of in-situ instance segmentation of on-branch soybean pods with different instance segmentation network and different transfer learning strategies are shown in Table \ref{tab:table 4} and Figure \ref{fig:10} respectively. As shown in Table \ref{tab:table 4}, our proposed method outperforms the baseline methods, which is in line with the results shown in Figure \ref{fig:10}, where the proposed two-step transfer learning method achieves more accurate segmentation results. Thus, the results demonstrate that the synthetic in-vitro soybean pods dataset can be the supplement of mature soybean plant dataset for in-situ segmentation of on-branch soybean pods. More evaluation results can be seen in the supplemental material. The comparative experiments reveal that the model which was fine-tuned on the synthetic in-vitro soybean pod dataset only can learn the pods region roughly. However, if the model transfer learning from synthetic in-vitro soybean pod dataset in the first step, and then fine-tuning on the real-world soybean plant dataset, the performance will be improved compared with directly finetuned on real-world soybean plant dataset.


 \begin{table*}[htp]
\centering
 \caption{ Evaluation on mature soybean plant test dataset with different instance segmentation models and transfer learning strategies.}
\begin{tabular}{ccccc}
\toprule[1.5pt] \text { Transfer learning strategy } & \text { Recall@[.5:.95] } & \text { AP${_{50}}$ } & \text {  AP${_{75}}$ } & \text{  AP@[.5:.95] } \\
\midrule[0.5pt] 
\text{Yolact+(COCO,Dataset\_in-vitro) } & 0.042 & 0.043 & 0.019 & 0.021 \\
\text{Yolact+(COCO,Dataset\_on-branch)} & 0.319 & 0.474 &0.079 & 0.168 \\
\text{Yolact+ours} & 0.419 & 0.579 &0.219 & 0.269 \\
\hline
\text{Mask RCNN+(COCO,Dataset\_in-vitro) } & 0.052 & 0.043 &0.021 & 0.022 \\
\text{Mask RCNN+(COCO,Dataset\_on-branch) } & 0.547 & 0.687 &0.379 & 0.392 \\
\text{Mask RCNN+ours} & 0.555 & 0.694 &0.423 & 0.398 \\
\hline
\text{Blendmask+(COCO,Dataset\_in-vitro)} & 0.260 & 0.037 &0.009 & 0.015 \\
\text{Blendmask+(COCO,Dataset\_on-branch)} & 0.543 & 0.656 &0.345 & 0.352 \\
\text{Blendmask+ours} & 0.556 & 0.686 &0.345 & 0.352 \\
\hline
\text{Swin transformer+(COCO,Dataset\_in-vitro)} & 0.125 & 0.100 &0.037 & 0.045 \\
\text{Swin transformer+(COCO,Dataset\_on-branch)} & 0.535 & 0.773 &0.519 & 0.477 \\
\text{Swin transformer+ours} & \color{red}{\textbf{0.575}} &  \color{red}{\textbf{0.80}}  &  \color{red}{\textbf{0.579}}  & \color{red}{\textbf{0.522}}  \\
  
\toprule[1.5pt]
\end{tabular}
\label{tab:table 4}
\end{table*}

\section{Conclusion}
The proposed method can be employed in automatic pods counting after instance segmentation of on-branch soybean pods. It solves the high cost and low efficiency problems of traditional artificial pods counting such as poor accuracy, and provides a richer and more accurate
phenotypic parameter reference for breeding experts to carry out the breeding design. Moreover, farmers can utilize it to estimate yields of soybean quickly.

The major contribution and advantages of our method are: (1) A novel synthetic image generation method is proposed for automatically creating labeled high throughput in-vitro soybean pods image set. (2) A new hybrid sim/real and in-vitro/on-branch dataset, including synthetic invitro soybean pods dataset and real-world mature soybean plant dataset, designed for transferring from simulation to reality and from in-vitro segmentation to on-branch segmentation robustly. (3) The proposed two-step transfer learning method on a tiny Swin transformer based instance segmentation network achieves a decent performance for in-situ segmentation of on-branch soybean pods, which is the first work utilizing a synthetic high throughput in-vitro soybean pods images dataset for mimicking on-branch soybean pods of the mature soybean plant which are of complex architecture with pods frequently touching each other.

However, the 2D image based method does not carry the depth (range) information about the plant leading a poor segmentation performance for badly occluded pods. 

In the future, we plan to similarly analyze in-situ segmentation and seed-per-pod estimation of on-branch soybean pod for mature soybean plant phenotype investigation. Another avenue for future work is incorporating spatial information into the analysis to improve segmentation accuracy.

\section{Acknowledgement}
This work was supported jointly by National Natural Science Foundation of China (31971786), the China Agriculture Research System(CARS-04), the National Key R\&D Program of China
(2017YFD0101400) and the Chinese Universities Scientific Fund(2021TC111). All of the mentioned support is gratefully acknowledged. In addition, thanks for all the help of the teachers and students of the related universities.

{\small
\bibliographystyle{ieee_fullname}
\bibliography{mybib}

\begin{thebibliography}{10}\itemsep=-1pt

\bibitem{baek2020high}
JeongHo Baek, Eungyeong Lee, Nyunhee Kim, Song~Lim Kim, Inchan Choi, Hyeonso
  Ji, Yong~Suk Chung, Man-Soo Choi, Jung-Kyung Moon, and Kyung-Hwan Kim.
\newblock High throughput phenotyping for various traits on soybean seeds using
  image analysis.
\newblock {\em Sensors}, 20(1):248, 2020.

\bibitem{barth2018data}
Ruud Barth, Joris IJsselmuiden, Jochen Hemming, and Eldert~J Van~Henten.
\newblock Data synthesis methods for semantic segmentation in agriculture: A
  capsicum annuum dataset.
\newblock {\em Computers and electronics in agriculture}, 144:284--296, 2018.

\bibitem{bolya2019yolact}
Daniel Bolya, Chong Zhou, Fanyi Xiao, and Yong~Jae Lee.
\newblock Yolact: Real-time instance segmentation.
\newblock In {\em Proceedings of the IEEE/CVF international conference on
  computer vision}, pages 9157--9166, 2019.

\bibitem{chandra2020active}
Akshay~L Chandra, Sai~Vikas Desai, Vineeth~N Balasubramanian, Seishi Ninomiya,
  and Wei Guo.
\newblock Active learning with point supervision for cost-effective panicle
  detection in cereal crops.
\newblock {\em Plant Methods}, 16(1):1--16, 2020.

\bibitem{chang2021exploring}
Fangguo Chang, Wenhuan Lv, Peiyun Lv, Yuntao Xiao, Wenliang Yan, Shu Chen,
  Lingyi Zheng, Ping Xie, Ling Wang, Benjamin Karikari, et~al.
\newblock Exploring genetic architecture for pod-related traits in soybean
  using image-based phenotyping.
\newblock {\em Molecular Breeding}, 41(4):1--21, 2021.

\bibitem{chen2020blendmask}
Hao Chen, Kunyang Sun, Zhi Tian, Chunhua Shen, Yongming Huang, and Youliang
  Yan.
\newblock Blendmask: Top-down meets bottom-up for instance segmentation.
\newblock In {\em Proceedings of the IEEE/CVF conference on computer vision and
  pattern recognition}, pages 8573--8581, 2020.

\bibitem{danielczuk2019segmenting}
Michael Danielczuk, Matthew Matl, Saurabh Gupta, Andrew Li, Andrew Lee, Jeffrey
  Mahler, and Ken Goldberg.
\newblock Segmenting unknown 3d objects from real depth images using mask r-cnn
  trained on synthetic data.
\newblock In {\em 2019 International Conference on Robotics and Automation
  (ICRA)}, pages 7283--7290. IEEE, 2019.

\bibitem{deng2009imagenet}
Jia Deng, Wei Dong, Richard Socher, Li-Jia Li, Kai Li, and Li Fei-Fei.
\newblock Imagenet: A large-scale hierarchical image database.
\newblock In {\em 2009 IEEE conference on computer vision and pattern
  recognition}, pages 248--255. Ieee, 2009.

\bibitem{desai2019automatic}
Sai~Vikas Desai, Vineeth~N Balasubramanian, Tokihiro Fukatsu, Seishi Ninomiya,
  and Wei Guo.
\newblock Automatic estimation of heading date of paddy rice using deep
  learning.
\newblock {\em Plant Methods}, 15(1):1--11, 2019.

\bibitem{dwibedi2017cut}
Debidatta Dwibedi, Ishan Misra, and Martial Hebert.
\newblock Cut, paste and learn: Surprisingly easy synthesis for instance
  detection.
\newblock In {\em Proceedings of the IEEE international conference on computer
  vision}, pages 1301--1310, 2017.

\bibitem{fehr1988principles}
Walter~R Fehr and James~R Justin.
\newblock Principles of cultivar development, vol. 2, crop species.
\newblock {\em Soil Science}, 145(5):390, 1988.

\bibitem{ghosal2019weakly}
Sambuddha Ghosal, Bangyou Zheng, Scott~C Chapman, Andries~B Potgieter, David~R
  Jordan, Xuemin Wang, Asheesh~K Singh, Arti Singh, Masayuki Hirafuji, Seishi
  Ninomiya, et~al.
\newblock A weakly supervised deep learning framework for sorghum head
  detection and counting.
\newblock {\em Plant Phenomics}, 2019, 2019.

\bibitem{groves2010estimating}
Frank~E Groves, Freddie~M Bourland, et~al.
\newblock Estimating seed surface area of cottonseed.
\newblock {\em J. Cotton Sci}, 14:74--81, 2010.

\bibitem{he2017mask}
Kaiming He, Georgia Gkioxari, Piotr Doll{\'a}r, and Ross Girshick.
\newblock Mask r-cnn.
\newblock In {\em Proceedings of the IEEE international conference on computer
  vision}, pages 2961--2969, 2017.

\bibitem{igathinathane2008shape}
C Igathinathane, LO Pordesimo, EP Columbus, WD Batchelor, and SR Methuku.
\newblock Shape identification and particles size distribution from basic shape
  parameters using imagej.
\newblock {\em Computers and electronics in agriculture}, 63(2):168--182, 2008.

\bibitem{kamilaris2018deep}
Andreas Kamilaris and Francesc~X Prenafeta-Bold{\'u}.
\newblock Deep learning in agriculture: A survey.
\newblock {\em Computers and electronics in agriculture}, 147:70--90, 2018.

\bibitem{kuznichov2019data}
Dmitry Kuznichov, Alon Zvirin, Yaron Honen, and Ron Kimmel.
\newblock Data augmentation for leaf segmentation and counting tasks in rosette
  plants.
\newblock In {\em Proceedings of the IEEE/CVF conference on computer vision and
  pattern recognition workshops}, pages 0--0, 2019.

\bibitem{lamprecht2007cellprofiler}
Michael~R Lamprecht, David~M Sabatini, and Anne~E Carpenter.
\newblock Cellprofiler™: free, versatile software for automated biological
  image analysis.
\newblock {\em Biotechniques}, 42(1):71--75, 2007.

\bibitem{lin2014microsoft}
Tsung-Yi Lin, Michael Maire, Serge Belongie, James Hays, Pietro Perona, Deva
  Ramanan, Piotr Doll{\'a}r, and C~Lawrence Zitnick.
\newblock Microsoft coco: Common objects in context.
\newblock In {\em European conference on computer vision}, pages 740--755.
  Springer, 2014.

\bibitem{liu2021swin}
Ze Liu, Yutong Lin, Yue Cao, Han Hu, Yixuan Wei, Zheng Zhang, Stephen Lin, and
  Baining Guo.
\newblock Swin transformer: Hierarchical vision transformer using shifted
  windows.
\newblock In {\em Proceedings of the IEEE/CVF International Conference on
  Computer Vision}, pages 10012--10022, 2021.

\bibitem{mcguire2021high}
Michael McGuire, Chinmay Soman, Brian Diers, and Girish Chowdhary.
\newblock High throughput soybean pod-counting with in-field robotic data
  collection and machine-vision based data analysis.
\newblock {\em arXiv preprint arXiv:2105.10568}, 2021.

\bibitem{nellithimaru2019rols}
Anjana~K Nellithimaru and George~A Kantor.
\newblock Rols: Robust object-level slam for grape counting.
\newblock In {\em Proceedings of the IEEE/CVF Conference on Computer Vision and
  Pattern Recognition Workshops}, pages 0--0, 2019.

\bibitem{pound2017deep}
Michael~P Pound, Jonathan~A Atkinson, Darren~M Wells, Tony~P Pridmore, and
  Andrew~P French.
\newblock Deep learning for multi-task plant phenotyping.
\newblock In {\em Proceedings of the IEEE International Conference on Computer
  Vision Workshops}, pages 2055--2063, 2017.

\bibitem{riera2020deep}
Luis~G Riera, Matthew~E Carroll, Zhisheng Zhang, Johnathon~M Shook, Sambuddha
  Ghosal, Tianshuang Gao, Arti Singh, Sourabh Bhattacharya, Baskar
  Ganapathysubramanian, Asheesh~K Singh, et~al.
\newblock Deep multi-view image fusion for soybean yield estimation in breeding
  applications deep multi-view image fusion for soybean yield estimation in
  breeding applications.
\newblock {\em arXiv preprint arXiv:2011.07118}, 2020.

\bibitem{sakurai2018two}
Shunsuke Sakurai, Hideaki Uchiyama, Atsushi Shimada, Daisaku Arita, and
  Rin-ichiro Taniguchi.
\newblock Two-step transfer learning for semantic plant segmentation.
\newblock In {\em ICPRAM}, pages 332--339, 2018.

\bibitem{tanabata2012smartgrain}
Takanari Tanabata, Taeko Shibaya, Kiyosumi Hori, Kaworu Ebana, and Masahiro
  Yano.
\newblock Smartgrain: high-throughput phenotyping software for measuring seed
  shape through image analysis.
\newblock {\em Plant physiology}, 160(4):1871--1880, 2012.

\bibitem{toda2020training}
Yosuke Toda, Fumio Okura, Jun Ito, Satoshi Okada, Toshinori Kinoshita, Hiroyuki
  Tsuji, and Daisuke Saisho.
\newblock Training instance segmentation neural network with synthetic datasets
  for crop seed phenotyping.
\newblock {\em Communications biology}, 3(1):1--12, 2020.

\bibitem{ubbens2018use}
Jordan Ubbens, Mikolaj Cieslak, Przemyslaw Prusinkiewicz, and Ian Stavness.
\newblock The use of plant models in deep learning: an application to leaf
  counting in rosette plants.
\newblock {\em Plant methods}, 14(1):1--10, 2018.

\bibitem{uzal2018seed}
Lucas~C Uzal, Guillermo~L Grinblat, Rafael Nam{\'\i}as, M{\'o}nica~G Larese,
  Julieta~Sofia Bianchi, Eligio~Natalio Morandi, and Pablo~M Granitto.
\newblock Seed-per-pod estimation for plant breeding using deep learning.
\newblock {\em Computers and electronics in agriculture}, 150:196--204, 2018.

\bibitem{xu2018srda}
Wenqiang Xu, Yonglu Li, and Cewu Lu.
\newblock Srda: Generating instance segmentation annotation via scanning,
  reasoning and domain adaptation.
\newblock In {\em Proceedings of the European Conference on Computer Vision
  (ECCV)}, pages 120--136, 2018.

\bibitem{yang2020efficient}
Si Yang, Lihua Zheng, Wanlin Gao, Bingbing Wang, Xia Hao, Jiaqi Mi, and Minjuan
  Wang.
\newblock An efficient processing approach for colored point cloud-based
  high-throughput seedling phenotyping.
\newblock {\em Remote Sensing}, 12(10):1540, 2020.

\bibitem{yang2022synthetic}
Si Yang, Lihua Zheng, Huijun Yang, Man Zhang, Tingting Wu, Shi Sun, Federico
  Tomasetto, and Minjuan Wang.
\newblock A synthetic datasets based instance segmentation network for
  high-throughput soybean pods phenotype investigation.
\newblock {\em Expert Systems with Applications}, 192:116403, 2022.

\bibitem{yu2019fruit}
Yang Yu, Kailiang Zhang, Li Yang, and Dongxing Zhang.
\newblock Fruit detection for strawberry harvesting robot in non-structural
  environment based on mask-rcnn.
\newblock {\em Computers and Electronics in Agriculture}, 163:104846, 2019.

\bibitem{zabawa2020counting}
Laura Zabawa, Anna Kicherer, Lasse Klingbeil, Reinhard T{\"o}pfer, Heiner
  Kuhlmann, and Ribana Roscher.
\newblock Counting of grapevine berries in images via semantic segmentation
  using convolutional neural networks.
\newblock {\em ISPRS Journal of Photogrammetry and Remote Sensing}, 164:73--83,
  2020.

\end{thebibliography}
}

\section{Appendix}
A. Addition results of our two-step transfer learning method with different instance segmentation networks

\begin{figure*}[t]
\begin{center}
\includegraphics[width=1\linewidth]{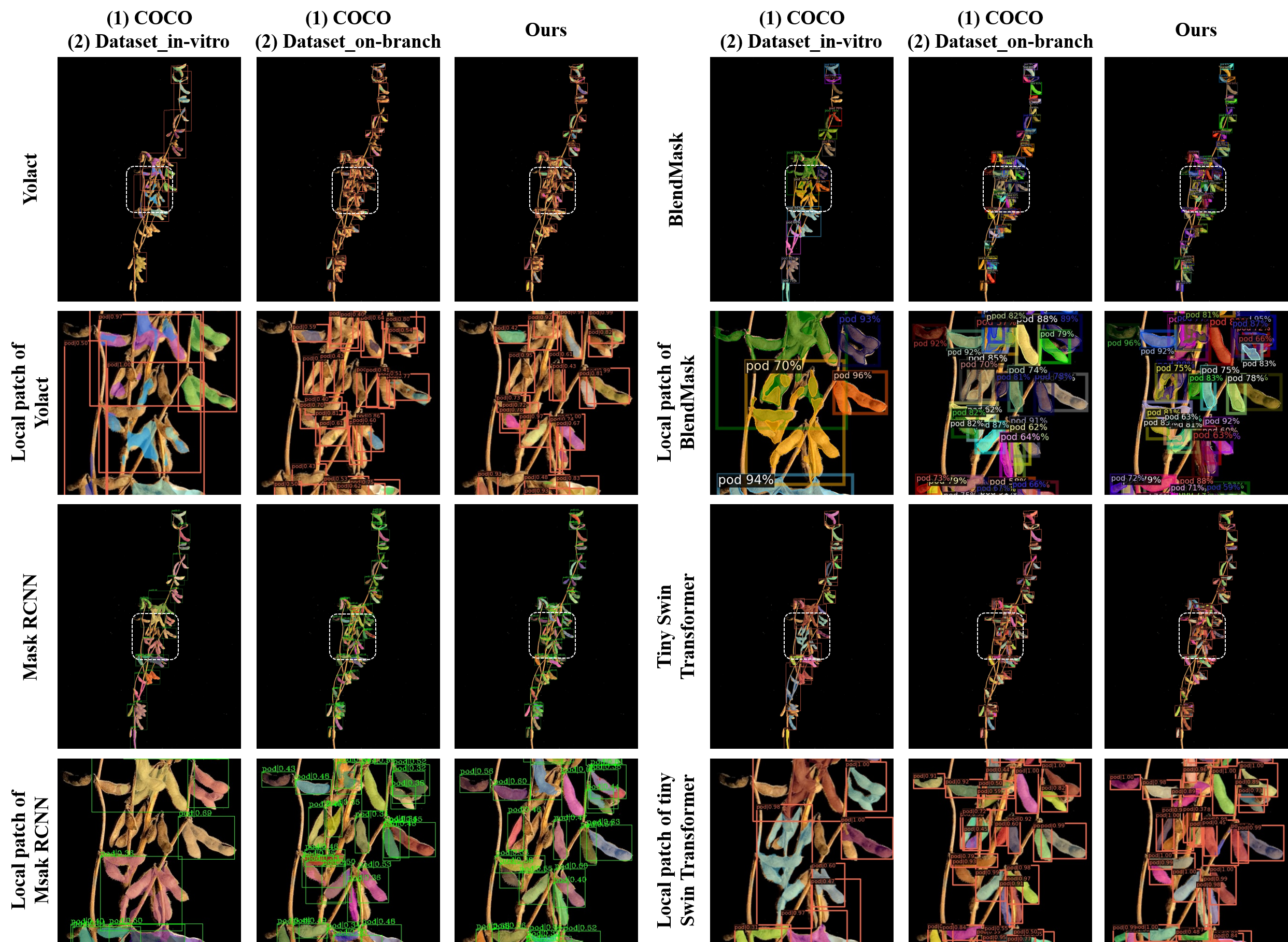}
\end{center}
\vspace{-0.5cm} 
     \caption{The effectiveness and efficiency of our two-step transfer learning method with different instance segmentation network}
\label{fig:11}
\end{figure*}

We conduct our two- step transfer learning method with different instance segmentation networks (Yolact \cite{bolya2019yolact}, Mask RCNN \cite{he2017mask},  BlendMask \cite{chen2020blendmask}, tiny Swin Transformer-based \cite{liu2021swin} instance segmentation network ) on our on-branch soybean pods phenotyping task. We also compare our two-step transfer learning method with the model purely trained by our synthetic in-vitro soybean pods dataset and the model only finetuned on a real world mature soybean plants dataset. The visualized results can be seen in Figure \ref{fig:11}. we can find that the performance of all the instance segmentation models can be improved observably by our two-step transfer learning method. And we can conclude that the model purely trained by our synthetic in-vitro soybean pods dataset can realize a very rough segmentation; the model only finetuned on a few real world mature soybean plants dataset can realize in-situ segmentation of on-branch soybean pods with low accuracy; the performance can be increased by our two-step transfer learning method which firstly trained by our synthetic dataset and then finetuned by a real world mature soybean plants dataset.

B. Addition results of different real world mature soybean plants

\begin{figure*}[t]
\begin{center}
\includegraphics[width=0.9\linewidth]{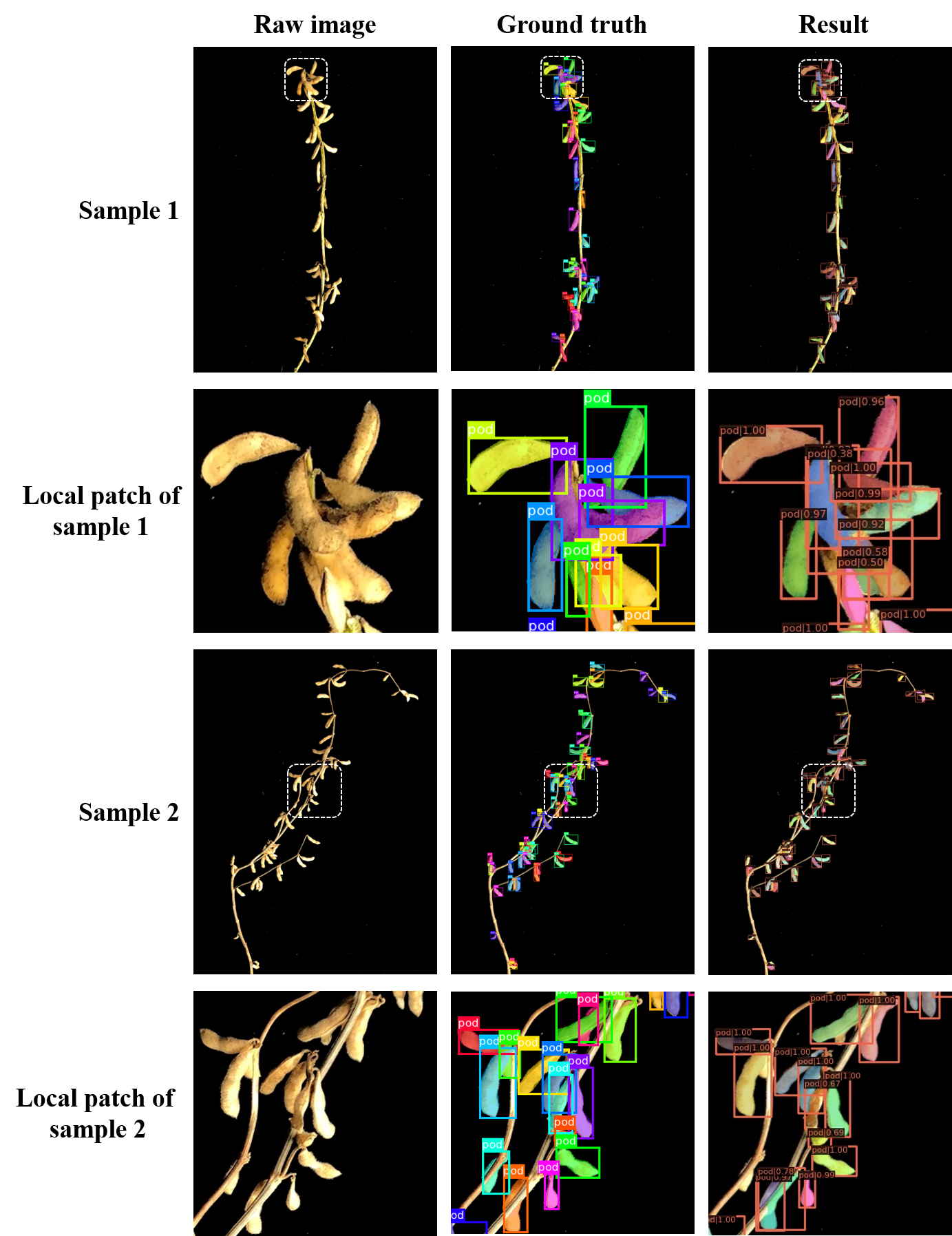}
\end{center}
\vspace{-0.5cm} 
   \caption{On-branch soybean pod in-suit instance result of different mature soybean plant sample in the test dataset with the best model}
\label{fig:12}
\end{figure*}

\begin{figure*}[t]
\begin{center}
\includegraphics[width=0.9\linewidth]{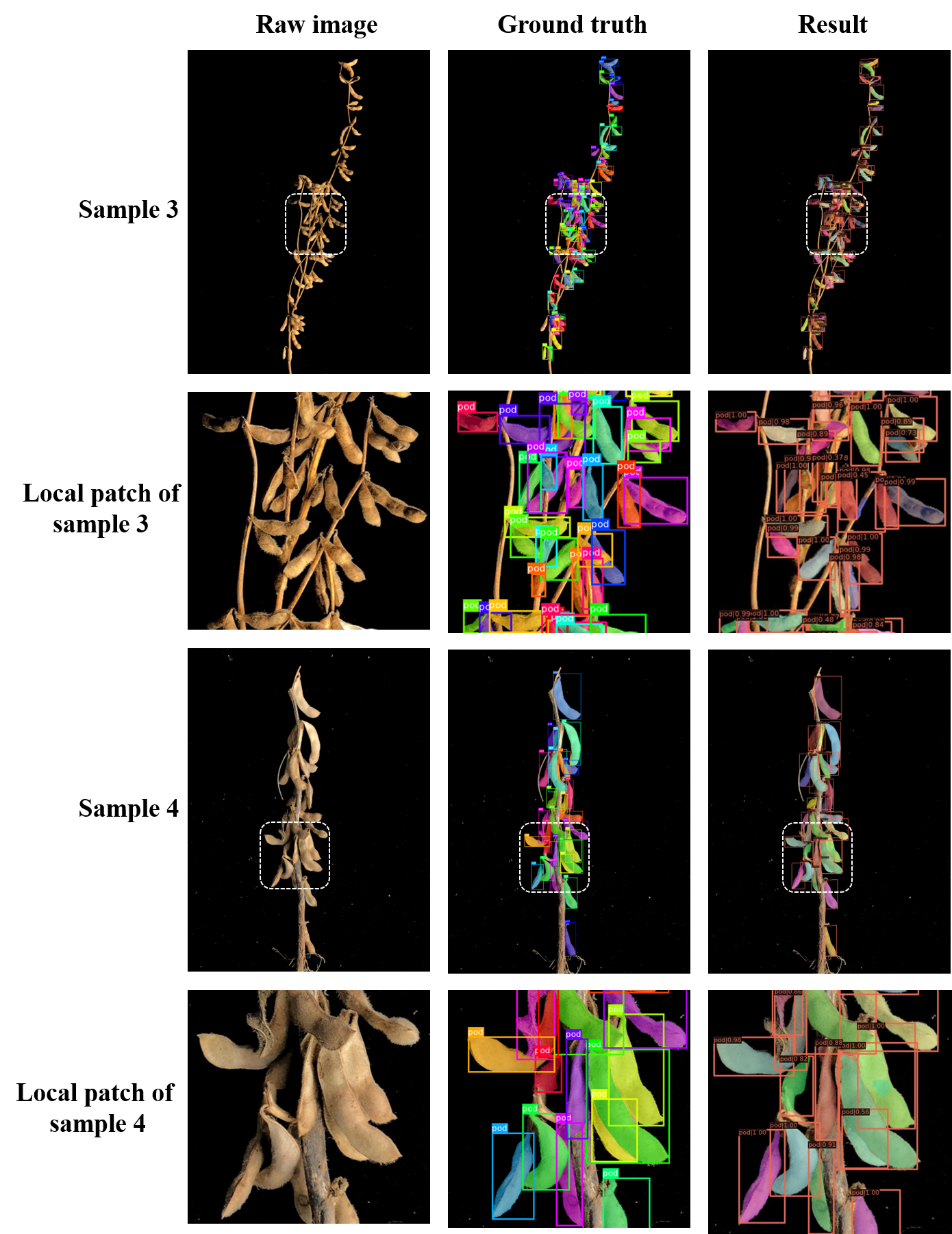}
\end{center}
\vspace{-0.5cm} 
     \caption{On-branch soybean pod in-suit instance result of different mature soybean plant sample in the test dataset with the best model}
\label{fig:13}
\end{figure*}

As shown in Figure \ref{fig:12} and Figure \ref{fig:13}, we report another on-branch soybean pod in-situ instance results of different mature soybean plant in the test dataset with the best model which finetuned the tiny Swin transformer based instance segmentation network with the 1600 synthetic in-virtu soybean pods images with 0.1 overlapping degree and a few samples of real world mature soybean plant images. These samples evaluation results of the best model as shown in Table \ref{tab:table 5}.

 \begin{table}
\centering
 \caption{ Some mature soybean plant samples evaluation results of the best model.}
\begin{tabular}{ccccc}
\toprule[1.5pt] \text { Sample } & \text { (1) } & \text { (2) } & \text { (3) } & \text{ (4) } \\
\midrule[0.5pt] \text{Recall@[.5;.95]} & 0.566 & 0.517 &0.596 & 0.577 \\
 \text{AP$_{50}$} & 0.924 & 0.779 & 0.876 & 0.777 \\
 \text{AP$_{75}$} & 0.637 &  0.581  &  0.689  & 0.680  \\
  \text{AP@[.5;.95]} & 0.763 &  0.618  &  0.686  & 0.698 \\
  
\toprule[1.5pt]
\end{tabular}
\label{tab:table 5}
\end{table}

\end{document}